\documentclass{article} % For LaTeX2e
\usepackage[preprint]{neurips_2024}

\usepackage{amsmath,amsfonts,bm}

\def\eqref#1{equation~\ref{#1}}
\def\1{\bm{1}}

\DeclareMathAlphabet{\mathsfit}{\encodingdefault}{\sfdefault}{m}{sl}
\SetMathAlphabet{\mathsfit}{bold}{\encodingdefault}{\sfdefault}{bx}{n}

\usepackage{hyperref}
\usepackage{url}
\usepackage[utf8]{inputenc} % allow utf-8 input
\usepackage[T1]{fontenc}    % use 8-bit T1 fonts
\usepackage{booktabs}       % professional-quality tables
\usepackage{amsfonts}       % blackboard math symbols
\usepackage{nicefrac}       % compact symbols for 1/2, etc.
\usepackage{microtype}      % microtypography
\usepackage{xcolor}         % colors
\usepackage{amsmath}
\usepackage{graphicx}       % for \includegraphics
\usepackage{subcaption}     % for subfigures
\usepackage{tikz}           % for tikzpicture overlays
\usepackage{algorithm}
\usepackage{algorithmic}
\usepackage{todonotes}      % for todo notes

\title{Divine Benevolence is an $x^2$: 
GLUs scale asymptotically faster than MLPs}

\author{Alejandro Francisco Queiruga\\
\texttt{alejandro.queiruga@gmail.com}
}

\makeatletter
\let\todonotes@originalcmd\todo
\newenvironment{todoblock}
  {\def\@todotext{}\begin{lrbox}{\@tempboxa}\begin{minipage}{\linewidth}}
  {\end{minipage}\end{lrbox}\todonotes@originalcmd[inline]{\usebox{\@tempboxa}}}
\makeatother
\let\todo\todoblock

\begin{document}

\maketitle

\begin{abstract}
Scaling laws can be understood from ground-up numerical analysis, where traditional function approximation theory can explain shifts in model architecture choices.
GLU variants now dominate frontier LLMs and similar outer-product architectures are prevalent in ranking models.
The success of these architectures has mostly been left as an empirical discovery.
In this paper, we apply the tools of numerical analysis to expose a key factor: these models have an $x^2$ which enables \emph{asymptotically} faster scaling than MLPs.
GLUs have piecewise quadratic functional forms that are sufficient to exhibit quadratic order of approximation.
Our key contribution is to demonstrate that the $L(P)$ scaling slope is $L(P)\propto P^{-3}$ for GLUs but only $L(P)=P^{-2}$ for MLPs on function reconstruction problems.
We provide a parameter construction and empirical verification of these slopes for 1D function approximation.
From the first principles we discover, we make one stride and propose the ``Gated Quadratic Unit'' which has an even steeper $L(P)$ slope than the GLU and MLP.
This opens the possibility of architecture design from first principles numerical theory to unlock superior scaling in large models.
Replication code is available at\\ \url{https://github.com/afqueiruga/divine_scaling}.
\end{abstract}

\section{Introduction}
In contemporary LLMs, variants of GLUs \citep{dauphin2017language} are now the norm (e.g., Gemma \citep{team2025gemma} and Qwen \citep{yang2025qwen3}) and SOTA recommendation and ranking models similarly incorporate outer-product architectures (e.g., Wukong \citep{zhang2024wukong} and Deep \& Cross Networks \citep{wang2017deep}).
% Contemporary model architecture development has thus for been empirical hill climbing on varying this feed forward element.
Famously, the success of GLUs was an empirical observation attributed to ``divine benevolence'' \citep{shazeer2020glu}.
% Famously, \cite{shazeer2020glu} closes with
% \begin{quote}
%    We offer no explanation as to why these
%    architectures seem to work; we attribute their success, as all else, to divine benevolence.
% \end{quote}
%  We make the argument that a key factor is that the $x^2$ terms greatly increase the expressivity on the model.
This work proposes a new understanding through a numerical function-approximation lens for the GLU's empirical success.
GLUs form piecewise quadratic functions, over MLPs' piecewise linear representation, which begets thinking along the lines of Taylor approximation.

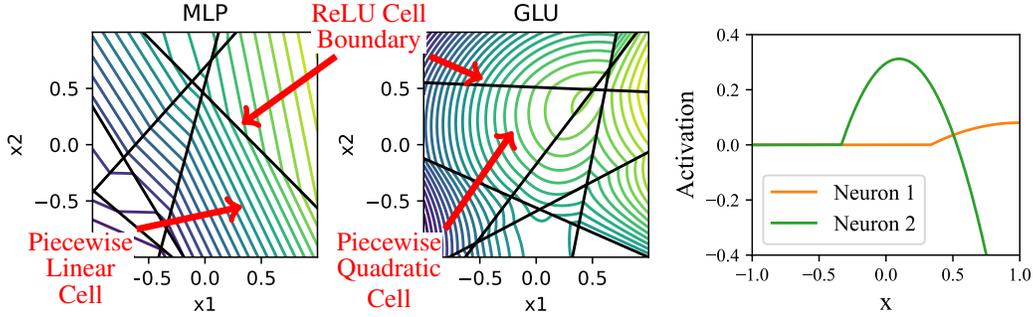
\begin{figure}
   \centering
   \includegraphics[width=3.5in]{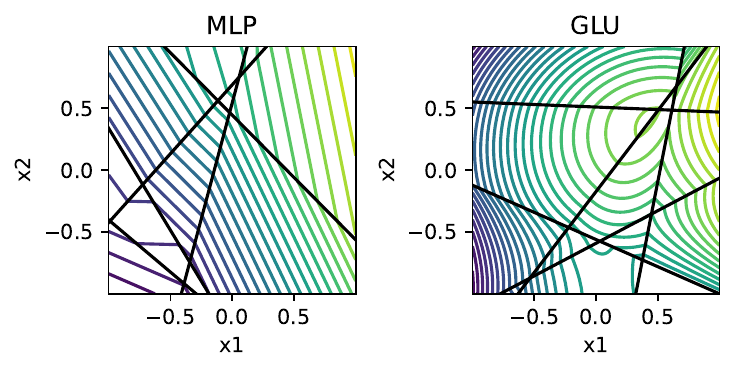}%
   \includegraphics[width=2in]{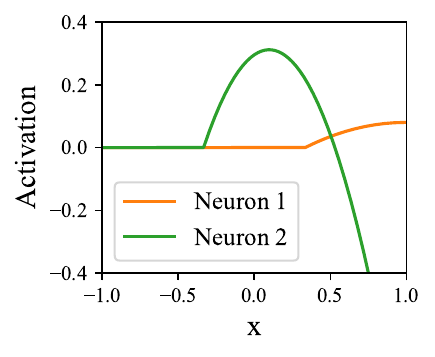}%
   \begin{tikzpicture}[overlay, remember picture, shift={( -4.2in,-0.15in )}]
      % Adjust (x,y) to appropriate coordinates relative to the left figure bounds.
      \draw[line width=2.5pt,->,red] (-1.4,1.7) node[below left,fill=white, inner sep=0pt] {\shortstack{Piecewise\\Linear\\Cell}} -- (0.0,2.0);
      \draw[line width=2.5pt,->,red] (2.7,1.7) node[below left,fill=white, inner sep=0pt] {\shortstack{Piecewise\\Quadratic\\Cell}} -- (3.6,3.0);
      \draw[line width=2.5pt,->,red] (2.5,4.0) node[above left,fill=white, inner sep=0pt] {\shortstack{ReLU Cell\\Boundary}} -- (3.2,3.7);
      \draw[line width=2.5pt,->,red] (1.25,4.0) -- (0.0,3.1);
   \end{tikzpicture}
   \caption{\label{fig:surface}Illustration of the function space of randomly initialized MLPs and GLUs with ReLU activations. Left and middle: randomly initialized MLP and GLU on $\mathbb{R}^2\rightarrow\mathbb{R}$. Black boundaries are the activation boundaries that break the domain into piecewise linear (MLP) or piecewise quadratic partitions. Right: in 1D, each neuron of the GLU forms a single piecewise quadratic function.}
\end{figure}

Scaling laws connect to the concept of convergence in numerical analysis.
In scientific software, the expectation of exact log-log slopes from derivation is even measured implementation validation (a ``convergence test'').
% (I.e., if an implementation has a slope of 2 when a 3 is expected, there is a bug in the code.)
When looking at a numerical method, an intuition is to look at the polynomial order of the method's underlying function approximation.
\cite{balestriero2020mad} used this insight to propose the Max Affine Spline Operator (MASO) interpretation of ReLU MLPs as piecewise linear splines.
% Each neuron (an index in the hidden state) potentially creates one piecewise linear region.
We build upon this interpretation to understand the GLU:
\begin{equation}
   GLU(x) = d + \sum_{i=0}^{n}D_i \text{relu}(G_ix+g_i)*(U_ix+u_i)
\end{equation}
and observe when the activation for the $i$-th neuron is "open", its contribution can be expanded as 
\begin{equation}
   \text{Active Neuron}_i (x) = D_i(G_iU_ix^2+(g_iU_i+G_iu_i)x+g_iu_i).
\end{equation}
% illustrated in 1D with a ReLU-like activation.
% From numerical analysis, function representations that include squared terms usually give cubic truncation error (converge in the number of parameters, not gradient descent iterations).
This makes it apparent that the GLU is a collection of quadratic basis functions with coefficients $D_i$. (We focus on ReLU activations, but GeGLUs and SwiGLUs also exhibit $\sigma(x)\rightarrow x$ upon activation.) Plotting randomly initialized networks in Fig.~\ref{fig:surface} makes this visually apparent.
% Following the numerical analysis intuition, we can demonstrate that the GLU can be asymptotically more accurate than a simple MLP.

% That is, for a simple MLP, doubling the number of parameters will halve the error.
% However, for a quadratic function, doubling the number of parameters will decrease the error by a factor of four.

To validate this intuition, we take the basis function approximation approach to NN error analysis and derive a scaling-optimal construction of network parameters in 1D, then empirically corroborate the expected scaling slopes for GLUs versus MLPs.
We find that for an MLP the error scales like $1/n^2$ (doubling parameters quarters the error), whereas for a GLU it scales like $1/n^3$ (doubling parameters reduces error by a factor of eight).

Deep networks are known (theoretically and empirically) to be more expressive.
From a component perspective, Shallow MLP and GLU are a standard building block in LLMs.
Studying shallow networks is a way of understanding the approximation capacity of a network component at building a small logical circuit.

\section{Background}

There has been extensive literature on understanding the accuracy of neural networks.
The empirical scaling law form made famous in \cite{hoffmann2022training} is:
\begin{equation}
L(D,P) = E + \frac{A}{P^\alpha} + \frac{B}{D^\beta}
\end{equation}
with parameter count $P$, training data size $D$, baseline error $E$, and curve fit parameters $A$, $\alpha$, $B$, and $\beta$.
(We will derive an $A$ and $\alpha$.)
Other forms have been proposed, e.g. \citep{li2025farseer}.
Scaling laws have been demonstrated in multiple domains, e.g. autonomous driving \citep{baniodeh2025scalinglawsmotionforecasting}.

While scaling behavior and architecture design is mostly an empirical art, there is extensive literature in approximation errors and scaling of NNs. Infinite limits of NNs have been analyzed theoretically \citep{neal1996priors,lee2017deep}. Derivations of approximation rates can be found in a few works in the literature:
\cite{barron1994approximation} proved a closed form error bound for sigmoid NNs as a function of width, $e = O(C_f^2/n) + O(n d_{dim} \log N_{data} / N_{data})$ for neurons $n$, dimensions $d_{dim}$, and number of training points $N$ when the target function has a minimum Fourier aspect length $C_f$, which also estimates $L(P)=1/P$ for sigmoid networks.
\cite{telgarsky2015representation} demonstrated constructions of deep ReLU networks were exponentially more expressive than shallow networks for a pathological discontinuous classification problem.
\cite{hanin2017approximating} provided error bounds for function reconstructions for deep relu networks.
The theoretical relu constructions of \cite{yarotsky2017error} included a lower bound of $L\propto 1/n^2$ for single layer relu networks.
\cite{bahri2024explaining} analytically determined four different regimes of scaling laws in large data, large parameter, and underparameterized regimes. They develop a $L(P)=\mathcal{O}(n^{-1})$ for linearized infinite width neural networks.

Recommendation systems motivated the outer product forms $y_k = W_{ijk} x_i x_j + b_k $ as "mixing" different features, e.g. explicitly crossing a document embedding against a user profile embedding \citep{wang2017deep,zhang2024wukong}.
While this crossing effect may also be a factor in GLUs, we show that even for \emph{scalar} problems the quadratic structure gives faster scaling.

\section{Theoretical Derivation of Scaling Slopes}

We derive an expectation for the MLP and the GLU on 1D function approximation for $L(P)$ in the limit of excessive data and training resources, $D\rightarrow\infty$.
The number of neurons is related to the parameter count by
$P_{MLP} = (dim_x + dim_y + 1) n + dim_y$
and
$P_{GLU} = (2 dim_x + dim_y + 2) n + dim_y$.
Let $f(x)$ denote the target function and $y(x)$ denote the NN approximation.

Consider a 1D dataset of points $\{x, f(x)\}$ with $x\in [0,1]$ where the goal is to reconstruct the function $f$ using a model $y$ using the Root Mean Squared Error (RMSE).
% \begin{equation}
%    X \sim \mathbb{U}[a,b] \quad Y = f(X)
%    \end{equation}
If the datapoints $x$ are sampled uniformly in the domain, in the infinite data limit, the error approaches an integral:
\begin{equation}
L = RMSE = \left(\frac{1}{|D|}\sum_{D} (f(x)-y(x))^2\right)^{1/2} \xrightarrow{|D|\to\infty} \left(\int_0^1(f(x)-y(x))^2\right)^{1/2}
\end{equation}
There is no noise or uncertainty such that the baseline error of the problem is $E=0$.
We provide the prior stated result by constructing a solution of parameters using the spline interpretation.
% \begin{equation}
% L_{MLP}(P) \propto \frac{1}{P^2}, \quad L_{GLU}(P) \propto \frac{1}{P^3}
% \end{equation}

\paragraph{Spline Partitioning using ReLU gates}
Firstly, we set the output bias to $d = f(0)$ arbitrarily.
We then utilize the ReLUs to construct equally sized partitions of size $h=1/n$ as follows: Set all gate boundaries to be $G_{i} = 1$ and the gate biases to be $g_i=-linspace(0,1,n)$.
This forms regularly spaced activations (see Fig.~\ref{fig:spline_inits}) that break the domain into $n-1$ closed partitions.
See the appendix for a full rollout; for the GLU, one partition has the form
\begin{equation}
   y_k(x) := d + \sum_{i=0}^{i<k} D_i(x - i h)(U_i x + u_i)\quad for\quad (k-1)h < x \leq kh
\end{equation}
% Each $y_k$ denotes the spline function within the cell with active neuron $k$.
% The ReLUs have been exchanged for the switch case of splines.
By construction, exactly one new neuron activates as we cross cell boundaries left-to-right.
This allows for solving parameters of individual neurons cell-by-cell using the following procedures.

\paragraph{MLP Parameter Construction} We solve for the weights $D_i$  such that at each segment, $y(ih) = f(ih)$ for all points $i=0,1,...N$. We do this moving from left to right. Within the leftmost segment, only one neuron is active, so we solve for the only unknown neuron $D_0 h + d = f(h)$, yielding
$
D_0 = (f(h) - f(0)) / h
$.
At the next partition cell, $D_1 = (f(2h) - f(0) - 2D_0h) / h$. The construction can be completed by solving cell-by-cell $y_i(ih)=f(ih)$ by a single pass recurrence relation
\begin{equation}
   D_i = (f(ih) - f(0))/h  - y_{i-1}(ih).
\end{equation}
We continue this procedure to the last spline segment. This construction forms an MLP that linearly interpolates at equally spaced node boundaries $x=ih$.
The local truncation error $\tau_i(x)=y_i(x)-f(x)$ is
\begin{equation}
   \tau_0(x) = \frac{x^{2} {f''_0}}{2} - \frac{h x {f''_0}}{2} + \frac{x^{3} {f'''_0}}{6} - \frac{h^{3} x {f''''_0}}{24} + \mathcal{O}(h^4 f'''_0)% + \frac{x^{4} {f''''_0}}{24}- \frac{h^{2} x {f'''_0}}{6}
\end{equation}
The total error $L$ can be estimated by integrating the truncation error of a representative cell $\bar{\tau}$,
\begin{equation}
   L = \left(\int_0^1 \tau(x)^2\mathrm{d}x\right)^{1/2} = \left(\sum_{i=0}^n \int_{(i-1)h}^{ih} \tau_i(x)^2\mathrm{d}x\right)^{1/2} \propto \left( \frac{1}{h} \int_{0}^{h} \bar{\tau}^2(x)\mathrm{d}x\right)^{1/2}
\end{equation}
Performing this integral using $\tau_0$ as the representative cell and keeping only the leading term yields
\begin{equation}
% \error(x) \propto  \frac{f''_{0} h^{2}}{12} + \frac{f'''_{0} h^{3}}{8} +\frac{23 f''''_{0} h^{4}}{240} + ...
L_{MLP} \propto \frac{h^2 C_2}{\sqrt{120}} + \mathcal{O}(h^3) \propto \frac{1}{P^2}
\end{equation}
where $C_2$ is a bound on $f''(x)$ arising from the $f''_0$ in $\bar{\tau}$.
As $h=1/n\propto 1/P$, we obtain $L(P)\propto 1/P^2$, and demonstrate MLP can have the same approximation order as a linear spline.

\paragraph{GLU Parameter Construction} The additional free parameters $U_k$ and $u_k$ within cell $k$ can eliminate an additional term in the local truncation error.
% \begin{equation}
% y_k(x) = d + \sum_{i=0}^k \left(D_i U_i \right) x^2  + \left(D_i U_i g_i + D_i u_i\right)x + \left( D_i g_i u_i \right)
% \end{equation}
The recurrent formula for the GLU splines in cell $i$ is
$y_i(x) = y_{i-1}(x) + \left(D_i U_i \right) x^2  + \left(D_i U_i g_i + D_i u_i\right)x + \left( D_i g_i u_i \right)$.
When lining up $y_i(ih)=f(ih)$, we solve for $u_i$ instead of $D_i$. For the first cell to solve $u_0$,
\begin{equation}
   f(h) = D_0 h (U_0 h + u_0) + d \rightarrow{} u_0 = \frac{- D_{0} U_{0} h^{2} - f{\left(0 \right)} + f{\left(h \right)}}{D_{0} h}
\end{equation}
which leaves $D_0$ and $U_0$ free. These are set by minimizing the truncation error within the cell,
$
\tau_0(x) = - D_{0} x \left(U_{0} x + u_{0}\right) - f{\left(0 \right)} + f{\left(x \right)}
$.
% As before, taylor expanding %yields
% \begin{equation}
% \tau_0(x) = - D_{0} x \left(U_{0} x + u_{0}\right) + x f'_0 + \frac{x^2f''_0}{2} + \frac{x^3f'''_0}{6} + \mathcal{O}(x^4)
% \end{equation}
Substituting $u_0$ and expanding $f(x)$ and $f(h)$ yields
\begin{equation}
% \tau_0(x) = D_{0} U_{0} h x - D_{0} U_{0} x^{2} - \frac{h^{3} x {f''''_0}}{24} - \frac{h^{2} x {f'''_0}}{6} - \frac{h x {f''_0}}{2} + \frac{x^{4} {f''''_0}}{24} + \frac{x^{3} {f'''_0}}{6} + \frac{x^{2} {f''_0}}{2}
\tau_0(x) = D_{0} U_{0} h x - D_{0} U_{0} x^{2} - \frac{h^{2} x {f'''_0}}{6} - \frac{h x {f''_0}}{2} + \frac{x^{3} {f'''_0}}{6} + \frac{x^{2} {f''_0}}{2} + \mathcal{O}(h^4 f''''_0)
\end{equation}
With the two degrees of freedom remaining, we are able to cancel out two terms here by setting
\begin{equation}
   U_0 = f''_0/2 D_0
\end{equation}
with one extra degree of freedom. The truncation error then becomes
\begin{equation}
   % \tau_0(x) = - \frac{h^{3} x {f''''_0}}{24} - \frac{h^{2} x {f'''_0}}{6}+ \frac{x^{4} {f''''_0}}{24} + \frac{x^{3} {f'''_0}}{6} 
   \tau_0(x) =  - \frac{h^{2} x {f'''_0}}{6} + \frac{x^{3} {f'''_0}}{6} +  \mathcal{O}(h^4 f''''_0)
\end{equation}
This forms another recurrence procedure for constructing the GLU parameters: sweep cell by cell by solving $y_i(ih)=y_{i-1}(ih)$ for $u_i$, and then solve for $U_i D_i$ that cancels out the $f''_0$ terms in the truncation error.
This systematically bounds the local truncation error by $\tau(x)=\mathcal{O}(h^3 f'''_0)$.
The global RMSE, again obtained as the square root of the square integral, yields
\begin{equation}
L_{GLU}(P) \propto \frac{h^3C_3}{\sqrt{945/2}} + \mathcal{O}(h^4) \propto \frac{1}{P^3}
\end{equation}
where $C_3$ is a bound on $f'''(x)$. Thus, we demonstrate that we can construct a parameterization for the GLU that has the same error rate as a piecewise quadratic spline, with a scaling law $L(P) \propto 1/P^3$.

% This construction is illustrated in Figure X for MLPs and GLUs.
% Further optimizing the parameters $G_i$ and $g$ will give a more accurate approximation by adjusting cell boundaries, but the upper bound of cubic scaling is guaranteed.

\section{Empirical Measurement of Scaling Slopes}

\begin{figure}
   \centering
   \begin{subfigure}[t]{0.32\linewidth}
      \centering
      \includegraphics[width=\linewidth,trim=0.1cm 0.4cm 0.2cm 0.2cm,clip]{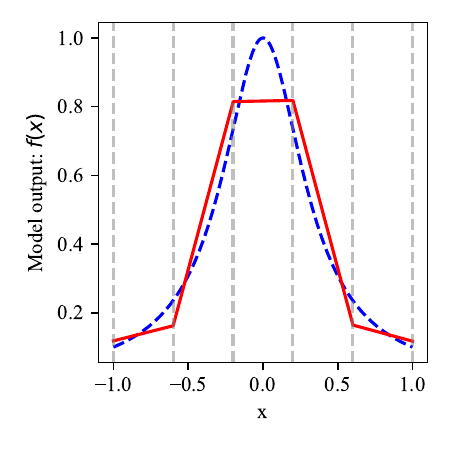}
      \caption{MLP: train only $\{D,d\}$ after spline initialization of $\{G,g\}$.}
   \end{subfigure}\hfill
   \begin{subfigure}[t]{0.32\linewidth}
      \centering
      \includegraphics[width=\linewidth,trim=0.1cm 0.4cm 0.2cm 0.2cm,clip]{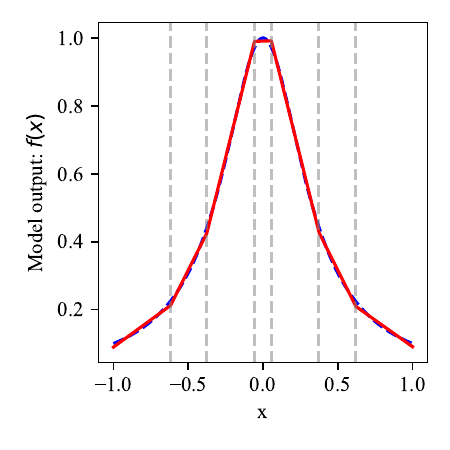}
      \caption{MLP: train all parameters.}
   \end{subfigure}\hfill
   \begin{subfigure}[t]{0.32\linewidth}
      \centering
      \includegraphics[width=\linewidth,trim=0.1cm 0.4cm 0.2cm 0.2cm,clip]{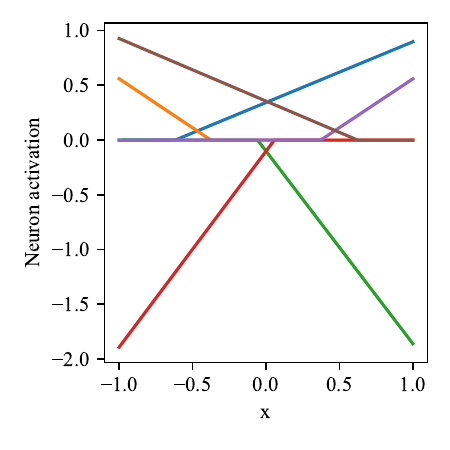}
      \caption{MLP neuron activations after training.}
   \end{subfigure}

   % \vspace{0.5em}

   \begin{subfigure}[t]{0.32\linewidth}
      \centering
      \includegraphics[width=\linewidth,trim=0.1cm 0.4cm 0.2cm 0.2cm,clip]{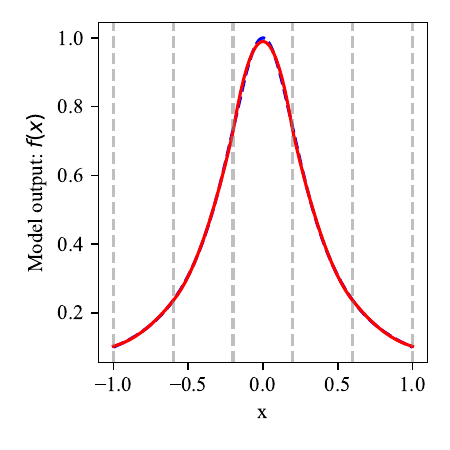}
      \caption{GLU: train only $\{D,d,U,u\}$ after spline initialization of $\{G,g\}$.}
   \end{subfigure}\hfill
   \begin{subfigure}[t]{0.32\linewidth}
      \centering
      \includegraphics[width=\linewidth,trim=0.1cm 0.4cm 0.2cm 0.2cm,clip]{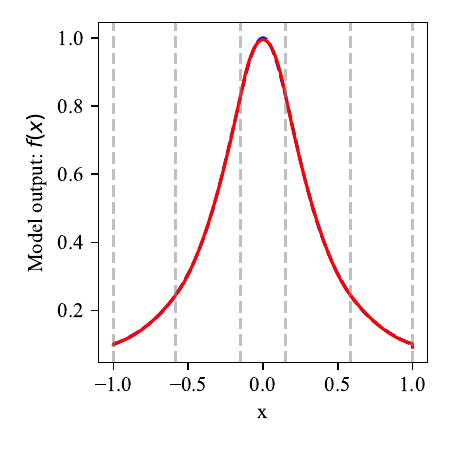}
      \caption{GLU: train all parameters.}
   \end{subfigure}\hfill
   \begin{subfigure}[t]{0.32\linewidth}
      \centering
      \includegraphics[width=\linewidth,trim=0.1cm 0.4cm 0.2cm 0.2cm,clip]{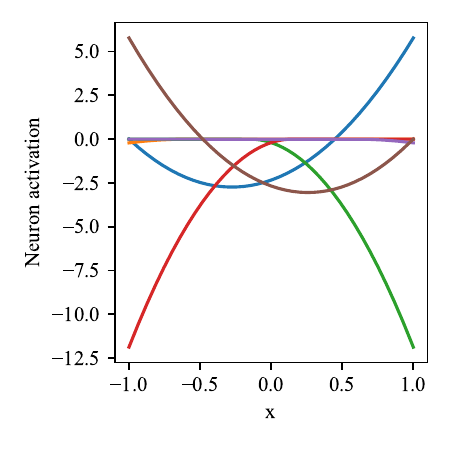}
      \caption{GLU neuron activations after training.}
   \end{subfigure}

   \caption{\label{fig:mlp_glu_solutions}Comparison of MLP and GLU fitting behavior under spline-based initialization (a,d) and full training (b, e).
   Vertical lines denote cell boundaries formed by $Gx+g=0$.
   Partial optimization in (a,d) already achieves the same asymptotic scaling rate as the analytical construction.
   The activations of individual neurons of the fully trained MLP and GLU are shown in (c,f).}
\end{figure}

We fit the target function
$f(x)= (1+\cos^2(\pi x))^{-1}$ on 10k points sampled on $[-1, 1]$. (The higher order Taylor terms of this function decay slowly.)
We utilize the spline-based initialization scheme, $G_i=\pm 1$, and $g_i=\pm linspace(-1,1,n)$, with alternating signs to avoid the dead neurons (see Fig.~\ref{fig:spline_inits} in the appendix.)
The other parameters are initialized with a normal distribution of $U_i,u_i,D_i,d_i\sim \mathcal{N}(0,1)$.
To reach the lowest possible error to extend the convergence plot as far as possible, the models are trained in double precision on CPU using Newton's method.
The optimization employs layer-by-layer full batch updates, Jacobi preconditioning, singular row elimination, and line search.
The PyTorch replication code is available at\\ \url{https://github.com/afqueiruga/divine_scaling}.
%We cascade Newton's method layer-by-layer backwards, optimizing $D$ then $U$ then $G$, and then the whole network.

% At training time, the model can do better with the ability to adjust $G$.
The optimization process is illustrated in Fig.~\ref{fig:mlp_glu_solutions} for a 6 neuron MLP and GLU to approximate the target function (blue line).
The analytical construction keeps $G$ frozen and evenly spaces the cells formed by the activation functions, equivalent to the leftmost panels in both rows.
In the construction, the function value at knot boundaries (vertical dashed lines) lie on the target function; with MSE optimization they do not.
The analytical solutions and those in (a,d) achieve the theoretical optimal scaling slopes, even without achieving the lowest possible MSE.
Fully training the network yields the approximation in the middle panels and the neuron activations in the right panels.

\begin{figure}
   \centering
\includegraphics[width=2.5in]{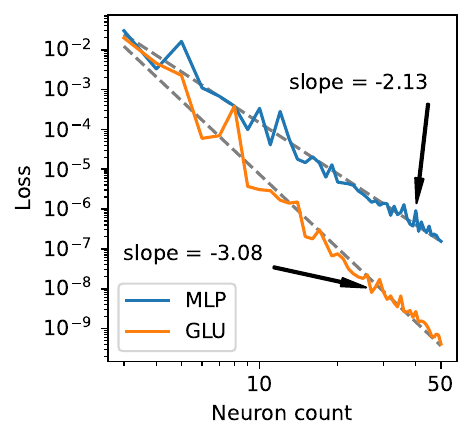}%
\includegraphics[width=2.5in]{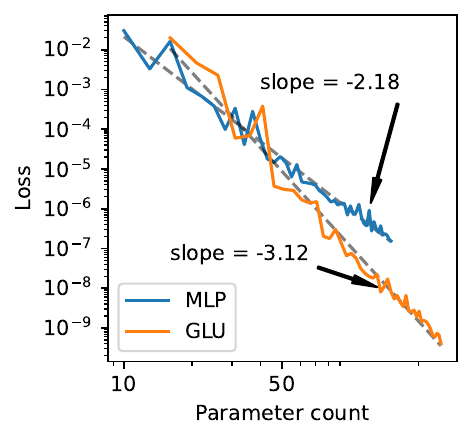}
   \caption{\label{fig:result1d}Experimental results for $L(P)$ scaling on 1d function approximation.
   The x-axis on the left uses neuron count which corresponds to the number of spline knots; changing the variable to parameter count on the right translates the log-log lines but does not change the log-log slope.}
\end{figure}

\begin{figure}
   \centering
   \includegraphics[width=2.5in]{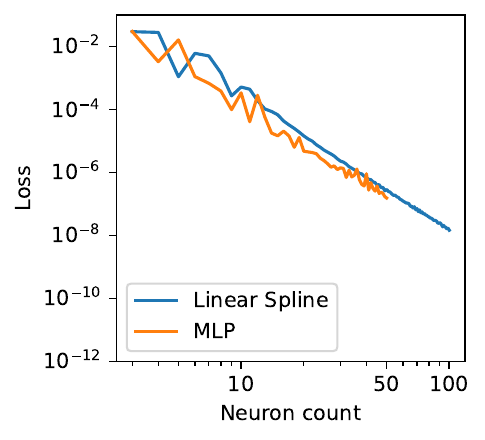}%
   \includegraphics[width=2.5in]{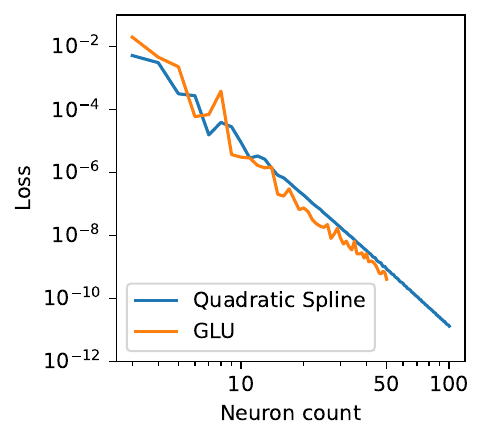}
   \caption{\label{fig:spline_comparison}MLPs and GLUs have similar approximation errors and order of convergence as their spline counterparts. Neuron count is analogous to the number of knots (control nodes) in a spline.}
\end{figure}

Fig.~\ref{fig:result1d} displays the results of the scaling estimation.
The NNs optimize on an MSE loss and report the Root Mean Square Error (RMSE) for each width starting from $n=1$ up to $n=50$.
Using RMSE instead of MSE matches local truncation order and numerical analysis theory.
Given all of these data points, we perform a log-log fit.
The actual errors are lower than the analytical construction, but bounded by the same rate.
Comparing the GLU to the MLP in the rightmost pane we see an \emph{asymptotic} reduction in error evidenced by different log-log slopes. The measured slopes match the analytical expectations:
\begin{itemize}
\item The MLP measures a slope of $n^{-2.13}$ and $P^{-2.18}$, matching the expected rate of -2.
\item The GLU measures a slope of $n^{-3.08}$ and $P^{-3.12}$, matching the expected rate of -3.
\end{itemize}

As a baseline, we compare the NNs to linear and quadratic splines, also implemented in the same PyTorch training script, in Fig.~\ref{fig:spline_comparison}.
We see that the error of the MLP and GLU is comparable to the spline counterparts.
The MLP and GLU achieve slightly smaller errors than their spline counterparts.
The additional degrees of freedom can move spline knots enabling a modest \emph{constant factor} reduction in error.

% \subsubsection*{Achieving low error}

% \begin{todo}
% Describe the Newtons method cascade.
% \begin{itemize}
%    \item MASO initialization scheme
%    \item Newton coordinate descent
   
% \end{itemize}
% \end{todo}

% \begin{todo}
% \subsubsection*{Spiral}
% Do dual spires.
% \end{todo}

\section{First principles architecture design for scaling: Gated Quadratic Unit}

\begin{figure}
   \centering
\includegraphics[width=2.5in]{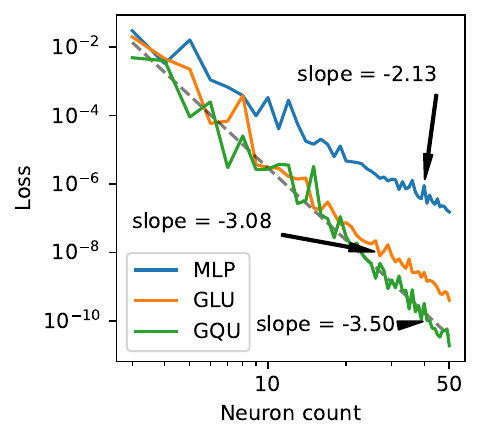}%
\includegraphics[width=2.5in]{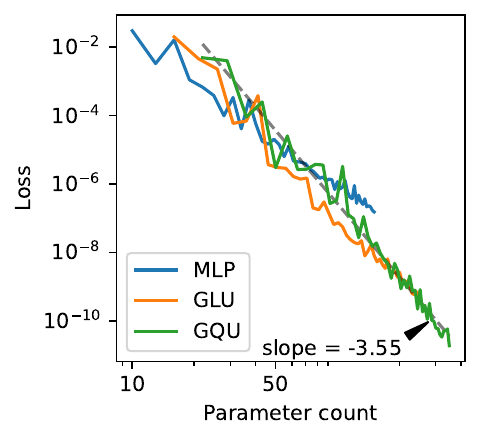}
   \caption{\label{fig:result1d_gqu}Novel architecture design to scale faster than a GLU: the Gated Quadratic Unit (GQU) has faster scaling than a GLU at $L(P)\propto P^{-3.5}$.}
\end{figure}

Given the confirmation that numerical analysis techniques apply to ML architectures, it is possible to construct a priori design architectures with faster scaling properties.
From first principles, we propose the Gated Quadratic Unit,
\begin{equation}
   GQU(x) = d + D(act(Gx+g)*(Ux+u)*(Qx+q))
\end{equation}
which has cubic terms in its unfolding.
Ideally, we expect it to be possible to have $L(P)\propto P^{-4}$. 
Repeating the empirical analysis from the main text, we observe a slope of -3.5 in Fig.~\ref{fig:result1d_gqu}.
This is lower than the gut instinct hypothesis, but still faster than the GLU, opening up the possibility for a new principled methodology for deriving new architectures.

\subsection{Higher dimensional synthetic and real problems}

\begin{figure}
   \centering
\includegraphics[width=2.5in]{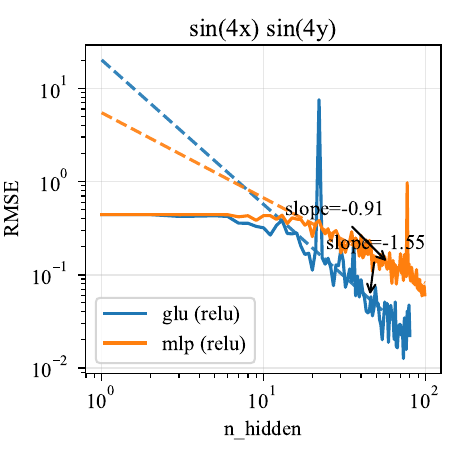}%
\includegraphics[width=2.5in]{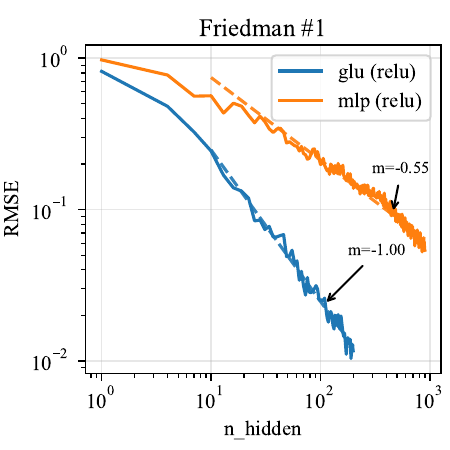}
\includegraphics[width=2.5in]{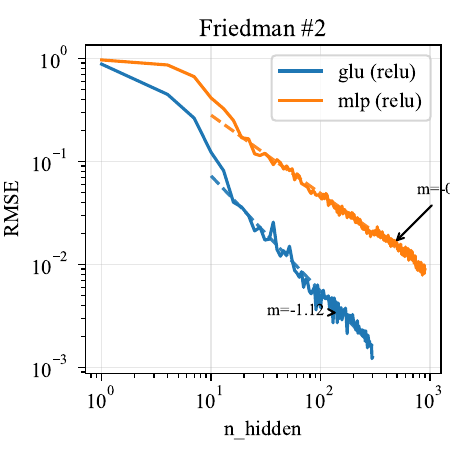}%
\includegraphics[width=2.5in]{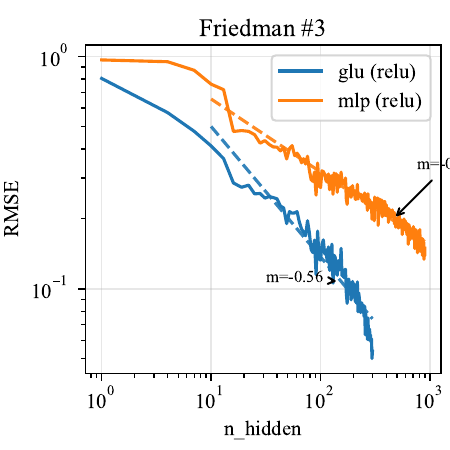}
   \caption{\label{fig:result_synth}Higher dimension synthetic problems.}
\end{figure}

\begin{figure}
   \centering
\includegraphics[width=2.5in]{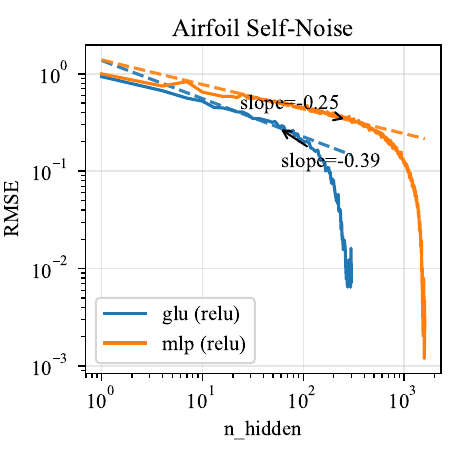}%
\includegraphics[width=2.5in]{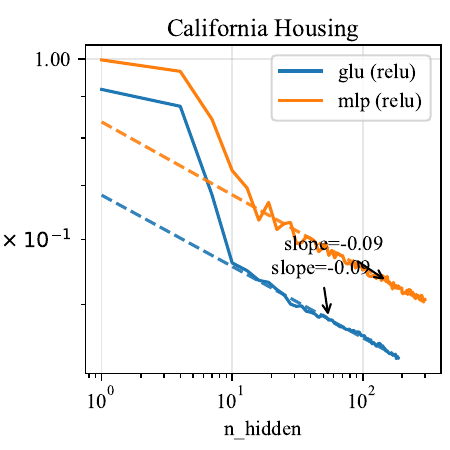}
   \caption{\label{fig:result_real}Real world benchmark regression datasets.}
\end{figure}

We empirically verify the different scaling slopes in higher dimensions for synthetic problems and real world regression datasets.
We use the Friedman synthetic problems from sklearn and two datasets from the UCI dataset, California housing and Airfoil Self-drag.
The results are plotted in figures \ref{fig:result_synth} and \ref{fig:result_real} and the measured slopes are summarized in table \ref{tab:slopetab}.
We observe that the scaling slopes for the GLU are steeper than the MLP for synthetic problems and real world datasets, except for the California housing dataset.
As expected, the scaling slope decreases with dimension by a factor of $1/dim_x$. However, not every problem matches the expected slope.

\begin{table}[h]
   \centering
   \caption{\label{tab:slopetab}Empirically measured slopes.}
   \begin{tabular}{|c|c|c|c|c|}
      \hline
      Problem & $x$ dim & MLP & GLU & GQU \\
      \hline
      $1/(1+cos(\pi x)^2)$ & 1 & $-2.13$ & $-3.08$ & -3.55 \\
      $sin(4x)sin(4y)$ & 2 & $-0.91$ & $-1.55$ & - \\
      Friedman1 & 5 & $-0.55$ & $-1.00$ & - \\
      Friedman2 & 4 & $-0.75$ & $-1.12$  & -\\
      Friedman3 & 4 & $-0.31$ & $-0.56$  & -\\
      Airfoil Self-drag & 5 & $-0.25$ & $-0.39$ & - \\
      California Housing & 8 & $-0.09$ & $-0.09$  & -\\
      \hline
   \end{tabular}
\end{table}

\section{Conclusion}

We systematically explained how to effect different $L(P)$ scaling slopes in NNs, which to our knowledge has not been published previously.
Using tools from classical numerical analysis, we show that higher-order approximation behavior appears in modern architectures, offering a lens for architecture design.
We posit that the superior order of approximation we demonstrated in GLUs may be a factor that caused them to win out against MLPs.
From first principles, we designed the GQU that has an even faster $L(P)$ scaling rate than the GLU and MLP.
% It is possible that different parts of a larger model might scale at different asymptotic rates.
We hypothesize that scaling large architectures may resemble mixed-method scientific simulations: doubling GLU width could require scaling other components at a different rate to maintain overall efficiency.
Our measurements are limited to slopes on 1D synthetic problems with shallow models.
% There are barriers such that this scaling effect may not be occurring in practice due to the following:
This effect may not dominate in large models on real-world datasets.
Higher-order scaling requires sufficiently smooth targets; real circuits may not be smooth enough to benefit.
The extra nonlinearities can also change optimization dynamics, preventing convergence to spline-optimal minima.
%  training to sufficient precision required layer-by-layer coordinate descent with Newton's method.
% Secondly, the higher order terms may not be necessary for the circuits learned by a neural network.
% However, we showed that GLU units can represent complex gates with asymptotically fewer hidden layers than MLPs.
This perspective still highlights \emph{efficient} approximation away from infinite width: GLUs can represent curved decision boundaries and complex circuits to the same error with 15 neurons versus 50.
% For complex circuits,
% We hypothesize that means that the probability of finding good circuit representation is much easier -- fewer neurons can form a lottery ticket of sufficient accuracy to bootstrap a  circuit.
% From a lottery ticket hypothesis perspective, this increases the probability of the optimization stumbling upon a useful circuit.
% That is, if it only takes 10 neurons to find stumble upon a sufficiently accurate representation of a rotation versus 100 neurons, the network can be more successful.
% This efficiency also means that 10x as many such circuits can also fit within the same hidden state budget.
Follow-up work should validate convergence rates on real-world datasets and generalize the construction to arbitrary dimensions.
We expect the curse of dimensionality to reduce the slope, but for the GLU slope to still be steeper than the MLP slope.
%We expect $L_{GLU}\propto P^{-3/}$.

\subsubsection*{Acknowledgments}
The author thanks David Hansul Park for discussing the initial idea and providing feedback on experiment design.

\bibliographystyle{iclr2026_conference}
\bibliography{refs}
\clearpage

\appendix

\section{Piecewise Spline Expansions}

\begin{figure}
   \centering
   \includegraphics[width=2.25in]{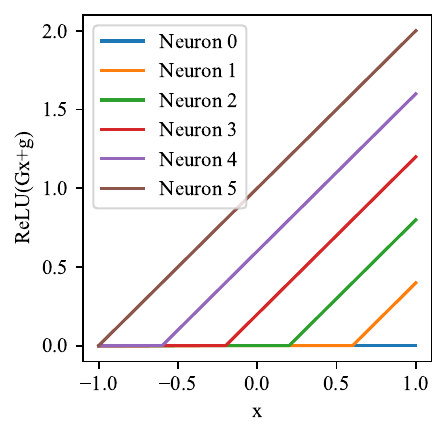}%
   \includegraphics[width=2.25in]{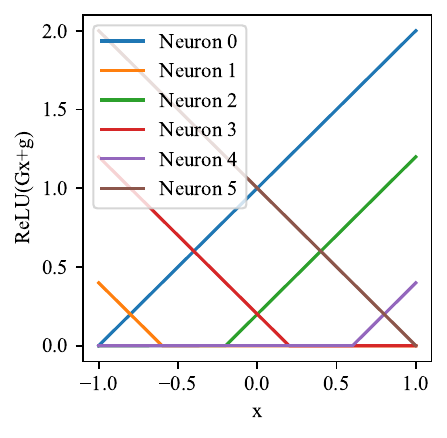}
   \caption{\label{fig:spline_inits}ReLU activations for the uniform cell spline initializations $G=\pm 1$ and $g=\pm linspace(-1,1,n)$. On the left is an initialization where all face the same way, which simplified the analytical construction. On the right is an alternating construction with the same partitioning but prevents all dead neurons.}
\end{figure}

Each cell references an overlapping set of parameters, as the neural network function approximation is not constructed as compact cells (as are splines or partition of unity methods), but the combination of all active neurons in a cell can be analyzed as a piecewise spline, albeit with an awkward parameterization.
The piecewise spline construction of the MLP that can be verbosely written out as a conditional, where each row corresponds to a different cell.

% \begin{align}
%    y(x) = \begin{cases}
%       d &;\quad x < 0 \\
%       d + D_0 x (U_0 x + u_0) &; \quad 0 < x < h \\
%       d + D_0 x (U_0 x + u_0) + D_1(x - h)(U_1 x + u_1) &; \quad h < x < 2h \\
%       ... \\
%       d + \sum_{i=0}^{i\neq1,i<n} D_i(x + i h)(U_i x + u_i) &; \quad (n-1)h < x < nh \\
%       % d + \sum_{i=0}^{i\neq1,i<N} D_i(x + i h)(U_i x + u_i) &; \quad  x > 1
%    \end{cases}
% \end{align}
\begin{align}
   y(x) = \begin{cases}
      y_{oob} := d &;\quad x \leq 0 \\
      y_{0} := d + D_0 x  &; \quad 0 < x \leq h \\
      y_1 := d + D_0 x+ D_1(x - h) &; \quad h < x \leq 2h \\
      ... \\
      y_k := d + \sum_{i=0}^{i\leq k} D_i(x - i h) &; \quad kh < x \leq (k+1)h \\
      ... \\
      y_n := d + \sum_{i=0}^{i<n} D_i(x - i h) &; \quad nh < x 
      % d + \sum_{i=0}^{i\neq1,i<N} D_i(x + i h)(U_i x + u_i) &; \quad  x > 1
   \end{cases}
\end{align}
$y_{oob}$ denotes when $x$ is "out of bounds" of all of the ReLUs and the function is piecewise constant.
The analytical construction in the main text solves these cells row by row for $D_i$.

Putting all gates facing in the same direction simplifies the presentation of the construction.
In practice the signs on some gates can be flipped to fully cover $\mathbb{R}$, which prevents the leftmost edge of the domain from becoming dead.
This construction and one equivalent alternative are plotted in Fig.~\ref{fig:spline_inits}.

For the GLU, the partitions are constructed the same, but the spline is piecewise quadratic:
\begin{align}
   y(x) = \begin{cases}
      y_{oob} := d &;\quad x \leq 0 \\
      y_{0} := d + D_0 x (U_0 x + u_0) &; \quad 0 < x \leq h \\
      y_1 := d + D_0 x (U_0 x + u_0) + D_1(x - h)(U_1 x + u_1) &; \quad h < x \leq 2h \\
      ... \\
      y_k := d + \sum_{i=0}^{i\leq k} D_i(x - i h)(U_i x + u_i) &; \quad kh < x \leq (k+1)h \\
      ... \\
      y := d + \sum_{i=0}^{i<n} D_i(x - i h)(U_i x + u_i) &; \quad nh < x 
      % d + \sum_{i=0}^{i\neq1,i<N} D_i(x + i h)(U_i x + u_i) &; \quad  x > 1
   \end{cases}
\end{align}
The analytical construction in the main text also solves these equations row by row.

% \end{document}

\section{Neuron visualizations of 2D functions}

\begin{figure}
   \centering
   \includegraphics[width=4in]{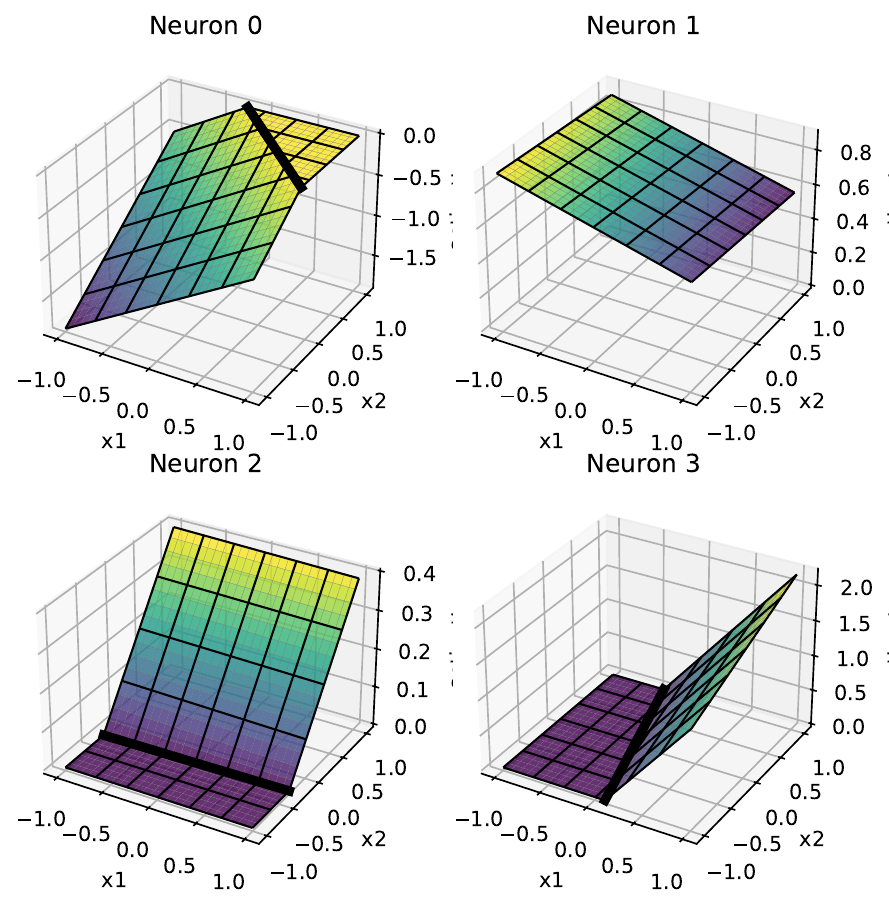}
   \caption{\label{fig:2d_mlp_neurons}Neuron activations of a randomly initialized MLP with four neurons form a piecewise linear basis set.}
\end{figure}

\begin{figure}
   \centering
   \includegraphics[width=4in]{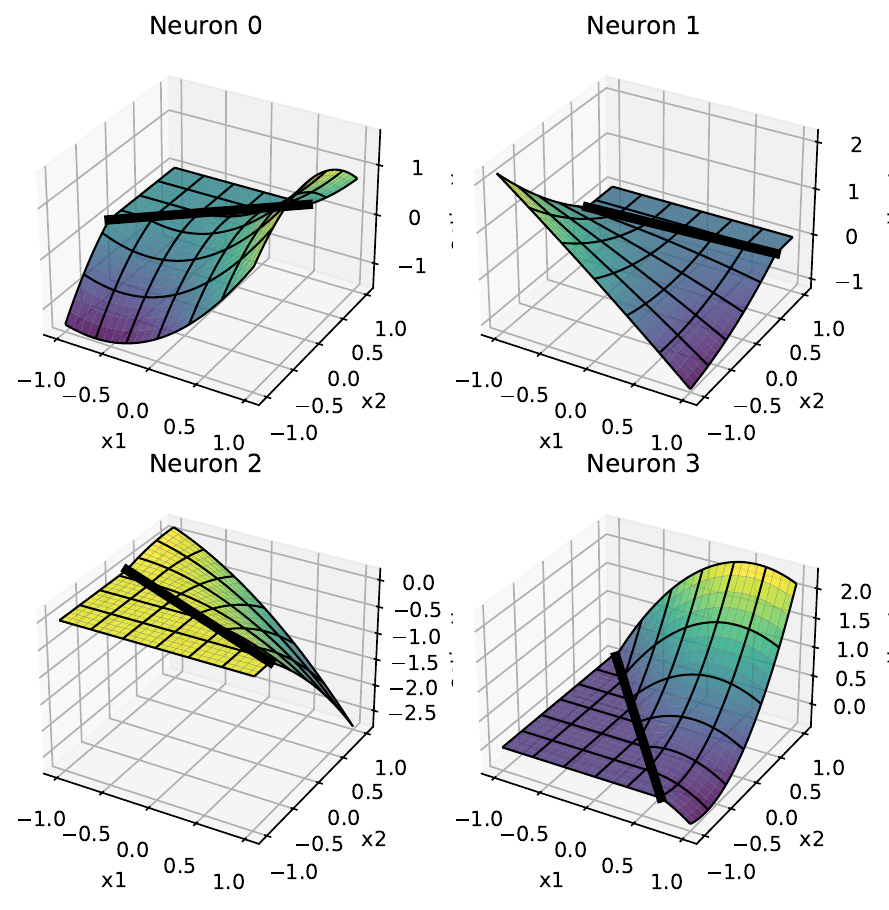}
   \caption{\label{fig:2d_glu_neurons}Neuron activations of a randomly  initialized GLU with four neurons form a piecewise quadratic basis set.}
\end{figure}

The partitioning into piecewise splines also applies to higher dimensions, where the activation functions form irregular cells along their hinges.
For any set of parameters, the boundary surfaces can be determined by solving for $G_{ij}x_j+g_i=0$ for each neuron $i$.
Randomly initialized $\mathbb{R}^2\rightarrow \mathbb{R}$ networks are shown in Fig.~\ref{fig:surface} with four neurons.
Each neuron's contributions are shown in Fig.~\ref{fig:2d_mlp_neurons} and Fig.~\ref{fig:2d_glu_neurons}, where the linear versus quadratic nature is apparent.

\begin{figure}
   \centering
   \includegraphics[width=4in]{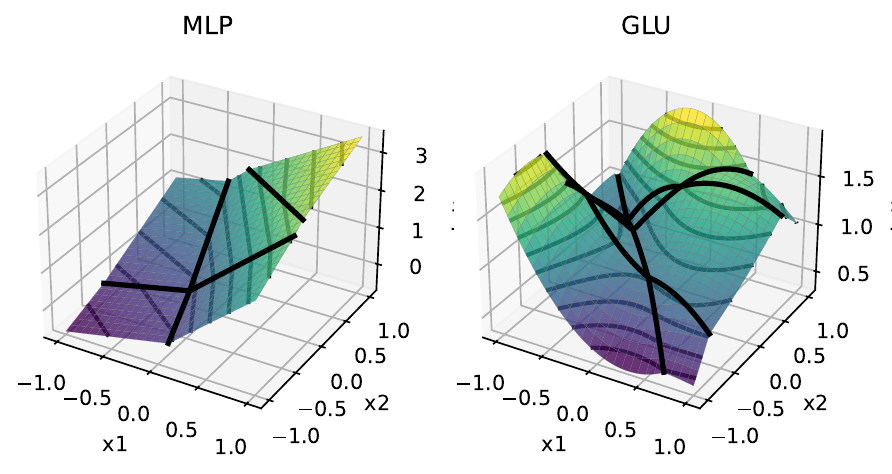}
   \caption{\label{fig:spline_surface}Surfaces of randomly initialized MLP and GLU with four neurons. The black lines demarcate the hinge of the ReLU activations, breaking the surface into piecewise linear and piecewise quadratic regions.}
\end{figure}

\end{document}